\documentclass[10pt, a4paper]{article}

\usepackage{lrec-coling2024} 

\usepackage{amsmath}
\usepackage{multirow}

\title{Class-Incremental Few-Shot Event Detection}

\name{Kailin Zhao$^{1,2}$, Xiaolong Jin$^{1,2*}$\thanks{{*}Corresponding author.}, Long
Bai$^{1}$, Jiafeng Guo$^{1,2}$, Xueqi Cheng$^{1,2}$} 

\address{$^{1}$Key Laboratory of Network Data Science and Technology, Institute
of\\
 Computing Technology, Chinese Academy of Sciences;\\
 $^{2}$School of Computer Science and Technology, University of Chinese
Academy of Sciences\\
\{zhaokailin17z, jinxiaolong, bailong18b, guojiafeng, cxq\}@ict.ac.cn}

\abstract{
Event detection is one of the fundamental tasks in information extraction and knowledge graph. However, a realistic event detection system often needs to deal with new event classes constantly. These new classes usually have only a few labeled instances as it is time-consuming and labor-intensive to annotate a large number of unlabeled instances.
Therefore, this paper proposes a new task, called class-incremental few-shot event detection. Nevertheless, this task faces two problems, i.e., old knowledge forgetting and new class overfitting. To solve these problems, this paper further presents a novel knowledge distillation and prompt learning based method, called Prompt-KD. 
Specifically, to handle the forgetting problem about old knowledge, Prompt-KD develops an attention based multi-teacher knowledge distillation framework, where the ancestor teacher model pre-trained on base classes is reused in all learning sessions, and the father teacher model derives the current student model via adaptation. 
On the other hand, in order to cope with the few-shot learning scenario and alleviate the corresponding new class overfitting problem, Prompt-KD is also equipped with
a prompt learning mechanism. Extensive experiments
on two benchmark datasets, i.e., FewEvent and MAVEN, demonstrate the
superior performance of Prompt-KD.
  \\ \newline  }

\begin{document}

\maketitleabstract

\section{Introduction}

Event detection is one of the fundamental tasks in information extraction and knowledge graph, which specifically extracts
trigger words from texts indicating the occurrence of events and further classifies
them into different event classes. For example, in ``\textit{Tom was
injured by falling rocks}'', the trigger word is \textit{``injured}'',
indicating an \textit{Injure} event. Event detection benefits
many downstream applications, e.g., event graph construction, question answering and information
retrieval.

A realistic event detection system often needs to deal with new classes of events, which continuously arrive.
Nevertheless, these new classes usually have only a few labeled instances as it is time-consuming
and labor-intensive to annotate a large number of unlabeled instances. Therefore, how
to incrementally learn the new event classes with only a few labeled instances has become
a challenging problem to the event detection system. To address this problem, in this paper we propose the Class-Incremental Few-Shot Event Detection (CIFSED) task.

Existing Few-Shot Event Detection (FSED) methods have achieved satisfying
performance~\cite{cong2021few,zhao2022knowledge}. Therefore, a straightforward
method for CIFSED is to train these FSED methods on base classes
and fine-tune them on new classes. However, if directly applying these methods 
in the CIFSED scenario via simple fine-tuning, two severe problems
will emerge~\cite{tao2020few}: 1) old knowledge forgetting: The model will forget old knowledge when dealing with new event classes and thus lower its own performance; 2) new class overfitting: The model is prone to overfitting to new classes and thus shows poor generalization
ability on subsequent classes, which is caused by the few-shot scenarios. Therefore, the primary objective of a CIFSED method is to learn new classes while maintaining old knowledge.

In order to achieve the similar objective in other fields such as image classification and named entity recognition, two kinds of Class-Incremental Few-Shot
Learning (CIFSL) methods have been proposed, i.e., topological structure based methods~\cite{tao2020topology,tao2020few}
and knowledge distillation based methods~\cite{cheraghian2021semantic,dong2021few,wang2022few}. The former kind of methods preserve the
old knowledge by maintaining the topology of the feature space in the network. The latter kind of methods maintain the output probabilities corresponding to the learned classes by adapting the model obtained from the last step based on new classes. Although this kind of methods have become the mainstream ones recently, they are defective in overcoming the old knowledge forgetting problem as this step-by-step manner causes the model to deviate from base knowledge as time goes by~\cite{dong2021few}. Furthermore, the new class overfitting problem has not been paid much attention~\cite{tao2020few,tao2020topology}.

To solve the above challenging problems, we propose a novel Knowledge
Distillation and Prompt learning based method, called Prompt-KD,
for CIFSED. Prompt-KD presents an attention based multi-teacher knowledge distillation framework. 
Therein, the ancestor teacher model, trained on base classes, is employed to derive the student model in the first learning session. 
From the second learning session onwards, the father teacher model, which is actually the student model in the last learning session, derives the new student model via adaptation. 
This framework also adopts an attention mechanism to balance the different importance
between these two teacher models as to the student model. 
To ease the forgetting problem about base knowledge, the ancestor teacher model is reused constantly in all learning sessions. 
Furthermore, Prompt-KD employs a prompt learning mechanism with additional
predefined texts (i.e., prompts) to the input instances in the support set, so as to cope with the few-shot learning scenario and alleviate the corresponding new class overfitting problem.

In summary, the main contributions of this paper are three-fold. 
\begin{itemize}
\item We propose for the first time, to the best of our knowledge, the 
Class-Incremental Few-Shot Event Detection (CIFSED) task, which often exists in the real-world event detection systems.
\item We propose a novel knowledge distillation and prompt learning based
method, called Prompt-KD, for CIFSED. To handle the forgetting problem about old knowledge, Prompt-KD presents an attention based multi-teacher
knowledge distillation framework. On the other hand, in order to  cope with the few-shot learning scenario and alleviate the corresponding new class overfitting problem,
Prompt-KD is also equipped with a prompt learning mechanism.
\item Extensive experiments on two benchmark datasets, i.e., FewEvent and
MAVEN, demonstrate the superior performance of Prompt-KD.
\end{itemize}

\section{Related Works}

\subsection{Class-incremental Few-shot Learning}

As aforesaid, there are two kind of approaches to the CIFSL task, i.e., topological
structure based and knowledge distillation based, respectively. Topological
structure based CIFSL methods preserve the
old knowledge by maintaining the topology of the feature space in the network. \citet{tao2020few} were the first
to propose the CIFSL task, and further presented a framework, called TOPIC, which
adopts a neural gas network to learn feature space topology for
knowledge representation. Next, \citet{tao2020topology} proposed
a new TPCIL framework, which employs an elastic Hebbian graph to
model the feature space topology. \citet{zhang2021few} adopted a decoupled
training strategy for representation learning and classifier learning
to ease the old knowledge forgetting problem. Knowledge
distillation based CIFSL methods maintains the output probabilities
corresponding to the learned classes. \citet{cheraghian2021semantic}
introduced semantic information into knowledge distillation and 
proposed a semantically-guided framework. Later, \citet{dong2021few}
put forward a relation knowledge distillation framework, which constrains
the relations among instances rather than their absolute positions. To address the CIFSED task,
we propose a novel attention based multi-teacher knowledge distillation
framework, which can repeatedly employ the ancestor teacher
model to handle the forgetting probl abouemt base knowledge.

\begin{figure*}[t]
\centering \includegraphics[width=0.9\textwidth]{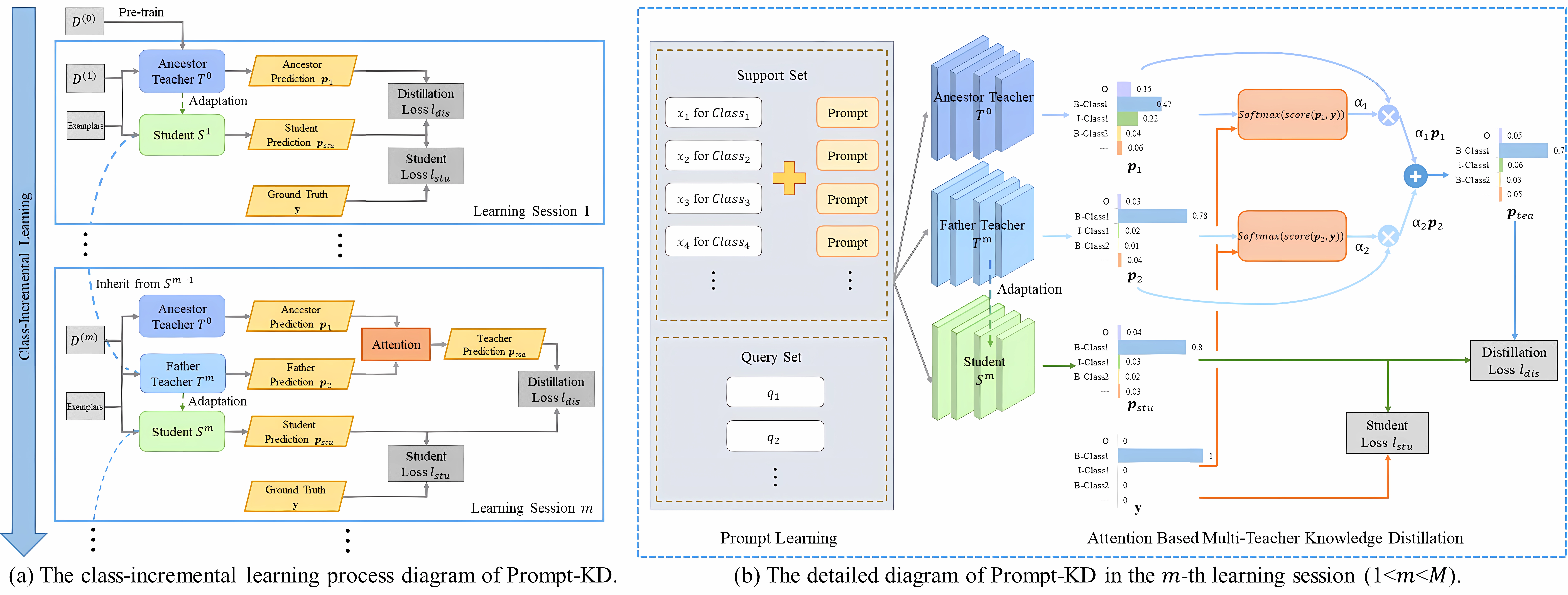}
\caption{The diagram of the Prompt-KD method.}
\label{fig:model}
\end{figure*}

\subsection{Event Detection}

There are two kinds of approaches to event detection, i.e., pipeline ones and joint
ones. Pipeline approaches follow the identification-then-classification
process and thus suffer from the error propagation problem. Due to
this reason, joint approaches have attracted much attention. Under the few-shot scenarios,
\citet{cong2021few} proposed PA-CRF based on a sequence tagging
method. Under the class-incremental scenarios, \citet{cao2020incremental}, \citet{yu2021lifelong} and \citet{liu2022incremental} solved event detection based on the knowledge distillation framework. 
In this paper, we choose the first and representative joint FSED model, i.e., 
PA-CRF, as our base model.

\subsection{Prompt Learning}

Prompt learning aims to minimize the gap between the pre-training
objective and the downstream fine-tuning objective. \citet{brown2020language}
proposed GPT-3, which is the first to employ prompts for downstream
tasks without introducing extra parameters, breaking the traditional
pre-training and fine-tuning mode. Later, prompt learning has
been applied in many information extraction tasks, e.g., entity extraction
\cite{cui2021template,liu2022qaner,ding2021prompt} and relation extraction
\cite{han2022ptr}. Recently, \citet{li2022piled} introduced prompt
learning into FSED and designed the cloze prompt as well as class-aware prompt
for event class identification and trigger localization, respectively.
In this paper, we design a new cloze prompt for joint FSED methods,
which can tackle the error propagation challenge.

\section{Problem Formulation}

Inspired by existing CIFSL works in other fields~\cite{dong2021few,tao2020few,wang2022few}, we formulate
CIFSED as follows. In CIFSED, we assume that a series of datasets $D^{(0)}$, $D^{(1)}$ ... $D^{(M)}$ constantly arrive, each of which is to be handled in a learning session. Here, $D^{(0)}$ is the large-scale dataset with base classes and $D^{(m)} (1 \leq m \leq M)$ is the $m$-th few-shot dataset containing new classes. All $D^{(m)} (0\leq m \leq M)$ have disjoint event class set with each other. Following the episodic learning strategy~\cite{laenen2021episodes}, there are several episodes in the 0-th learning session and only one episode in the $m$-th $(1 \leq m \leq M)$ learning session. For each episode, a support set and a query set are randomly instanced from $D^{(m)} (0\leq m \leq M)$, which is formulated in the $N$-way $K$-shot paradigm.
Given the support set
$S=\{(x_{i},y_{i})\}_{i=1}^{N\times K}$ which has $N$ classes and
each class has $K$ labeled instances, FSED aims to predict the labels
of tokens in the query set $Q$. In the support
set $S$, $x_{i}=\{w_{i}^{1},w_{i}^{2},...,w_{i}^{n}\}$ denotes an
n-word sequence, and $y_{i}=\{l_{i}^{1},l_{i}^{2},...,l_{i}^{n}\}$
denotes its corresponding label sequence of tokens. In the query set $Q=\{q_{i}\}_{i=1}^{N\times U}$, each class contains $U$ unlabeled instances,
where $q_{i}$ refers to a sequence of unlabeled tokens. 
Since joint FSED is formulated as a sequence tagging process, the label $l_i$ consists of two parts: the position part and the type part. For the position part, there are three types, i.e., B, I and O. B and I indicate that the corresponding word is the beginning and inside word of the event trigger, respectively, which may contain multiple words. O indicates that corresponding word does not belong to any trigger.
Therefore, the total
number of token labels is $2N+1$ ($N$ for \textit{B-Class},
another $N$ for \textit{I-Class}, and $1$ for label $O$).

\section{The Prompt-KD Method}

An illustrative diagram of the Prompt-KD method is presented in Figure~\ref{fig:model}, where the left part (a) illustrates the class-incremental learning process of Prompt-KD, while the right part (b) presents a more detailed implementation of the $m$-th learning session. As we can see, 
Prompt-KD consists of two main modules in each learning session, i.e., the attention
based multi-teacher knowledge distillation module and the prompt learning
module. The former module takes the prompt concatenated support instances and query instances as its inputs, and produces the prediction probabilities of the query tokens. The latter module
aims to add instructions to the support instances to help the training process and outputs
the enhanced instances with prompt.

\subsection{Attention Based Multi-Teacher Knowledge Distillation}

To address the old knowledge
forgetting problem, this module presents a two-teacher one-student knowledge distillation
framework and further employs an attention mechanism so as to
balance the different importance between these two teacher models.
This module contains five main components, i.e., the exemplars, the two teacher models, the attention
mechanism, the student model and the loss functions. The exemplars are selected in each learning session to replay the instances of old classes. They are taken by Prompt-KD as its input, together with the instances from new classes. The two teacher models produce their prediction probabilities respectively by inputting the prompt
enhanced support instances and query instances. The attention mechanism takes the above 
probabilities as its input and calculates the weighted teacher
probabilities. The student model takes the same input as the teacher models and outputs the student probabilities. The loss functions, including the distillation loss and the student loss, are employed to update the parameters of Prompt-KD.

\subsubsection{The Exemplars}

Generally, each learning session has its own exemplars, which are selected from the
learned classes~\cite{dong2021few}. For
example, in the $m$-th learning session $(m>0)$, the exemplars are obtained from the
instances in $D^{(0)}$, ..., $D^{(m-1)}$. Specifically, 
from $D^{(0)}$, a few randomly selected instances are adopted as the exemplars of base classes. 
Since all $D^{(i)}$ ($1\leq i\leq m-1$) have few-shot instances (e.g., 1-shot or 3-shot), they are all taken as the exemplars of the corresponding classes.
Then, in the $m$-th learning session, Prompt-KD takes $D^{(m)}$ together with the corresponding exemplars as its input, which contains a new support set $S'$ and a new query set $Q'$.

\subsubsection{The Two Teacher Models}

To overcome the old knowledge forgetting problem, Prompt-KD adopts two teacher models, i.e., the ancestor teacher model $T^{0}$ and the father teacher model $T^{m} (m>1)$.
The former refers to the model pre-trained
on the large-scale dataset $D^{(0)}$, whilst the latter is actually
the student model $S^{m-1} (m>1)$ in the last learning session, as shown in Figure~\ref{fig:model}(a). In this paper,
we adopt the pre-trained PA-CRF as the ancestor teacher model for
FSED, which mainly consists of four units, i.e., encoder unit,
emission unit, transition unit and decoder unit. 

The encoder unit takes the instances in the support set $S'$
and the query set $Q'$ as its input and maps them into the embedding space to represent their semantic
meanings. Given an input $x_{i}=\{w_{i}^{1}, w_{i}^{2}, ..., w_{i}^{n+P}\}$,
BERT-base-uncased~\cite{kenton2019bert} is employed to get its embeddings
as
$
\boldsymbol{x}_{i}=\{\boldsymbol{w}_{i}^{1},\boldsymbol{w}_{i}^{2},...,\boldsymbol{w}_{i}^{n+P}\}=BERT(x_{i}),
$
where $\boldsymbol{w}_{i}^{j}$ denotes the representation
of token $w_{i}^{j}$, which is of $H$ dimension, and $P$ is the length of the prompt. Thus, the support
instance embedding set $\boldsymbol{S}'$ can be formulated as
\begin{equation}
\boldsymbol{S}'=\{\boldsymbol{x}_{1},\boldsymbol{x}_{2},...,\boldsymbol{x}_{\left(m\times N+B\right)\times K}\},
\end{equation}
where $B$ is the number of event classes in $D^{(0)}$.

Similarly, the query instance embedding set $\boldsymbol{Q}'$ is formulated
as
\begin{equation}
\boldsymbol{Q}'=\{\boldsymbol{q}_{1},\boldsymbol{q}_{2},...,\boldsymbol{q}_{\text{\ensuremath{\left(m\times N+B\right)}}\text{\ensuremath{\times}}U}\},
\end{equation}
\noindent where $\boldsymbol{q}_{i}$ denotes the embedding representation
of $q_{i}$ by $\boldsymbol{q}_{i}=BERT(q_{i})$.

The emission unit takes the representations $\boldsymbol{S}'$ and $\boldsymbol{Q}'$ as its input and calculates the prototype $\boldsymbol{c}_{l}$ to each label $l$ of tokens based on $\boldsymbol{S}'$ as
\begin{equation}
\boldsymbol{c}_{l}=\frac{1}{|W(\boldsymbol{S}',{l})|}\sum_{\boldsymbol{w}\in W(\boldsymbol{S}',{l})}\boldsymbol{w},
\end{equation}
where $W(\boldsymbol{S}',{l})$ indicates the token set with label ${l}$
in $\boldsymbol{S}'$ and $\boldsymbol{w}$ is the representation of a token in it.
Then, this unit calculates the similarities between the presentations of the query
tokens and the prototypes as emission scores. In practice, the dot product operation is
chosen to measure the similarity.

The transition unit takes the representations of the prototypes as its input and then generates the parameters
(i.e., mean and variance) of Gaussian distribution as the transition scores.

Based on the
above obtained emission scores and transition
scores, the decoder unit calculates the probabilities
of possible label sequences for the given tokens in the query set $Q'$ and then derives the predicted label sequences. The Monte Carlo sampling
technique~\cite{gordon2019meta} is employed to approximate the integral.
In the inference phase, the first teacher model $T^{0}$ and the second one $T^{m}$ adopt the Viterbi algorithm
\cite{forney1973viterbi} to decode the probability distributions $\boldsymbol{p}_{1}$ and $\boldsymbol{p}_{2}$
to different label sequences for the query tokens, respectively. The
event detection process for the two teacher models can be simplified as 
\begin{equation}
\boldsymbol{p}_{1}=T^{0}({S'},{Q'}),\,\boldsymbol{p}_{2}=T^{m}({S'},{Q'}).
\end{equation}

\begin{table*}
\centering %
\begin{tabular}{|c|c|}
\hline 
{\small{}Stage 1} & {\small{}This is a {[}mask{]} event.{[}SEP{]} Its trigger words are
{[}mask{]}.}\tabularnewline
\hline 
{\small{}Stage 2} & {\small{}This is a {[}mask{]} event, which is learned {[}mask{*}{]}.{[}SEP{]}
Its trigger words are {[}mask{]}.}\tabularnewline
\hline 
{\small{}Stage 3} & {\small{}This is a {[}mask{]} event, which is learned {[}mask{*}{]}.{[}SEP{]}
Its trigger words are {[}mask{]}.}\tabularnewline
\hline 
\end{tabular}

\caption{The three-stage curriculum learning based prompts.}

\label{tab:cl-prompt}
\end{table*}

\subsubsection{The Attention Mechanism}

The attention mechanism is employed to balance the different importance
between the two teacher models. The final probability distribution of the teacher
models is denoted as $\boldsymbol{p}_{tea}$, which is calculated by
\begin{equation}
\boldsymbol{p}_{tea}=\sum_{i=1}^{2}{\alpha}_{i}\boldsymbol{p}_{i},
\end{equation}
where ${\alpha}_{i}$ refers to the weight of the $i$-th
teacher model. Moreover, ${\alpha}_{i}$ is obtained via ${\alpha}_{i}=Softmax(\boldsymbol{s}_{i})$.
Therein, $\boldsymbol{s}_{i}$ is calculated as
\begin{equation}
\boldsymbol{s}_{i}=score(\boldsymbol{p}_{i},\boldsymbol{y})=\boldsymbol{p}_{i}\boldsymbol{W}\boldsymbol{y},
\end{equation}
where $\boldsymbol{y}$ denotes the ground truth distribution of the
query tokens to different label sequences and $\boldsymbol{W}$ is a learnable matrix.

\subsubsection{The Student Model}

As shown in Figure~\ref{fig:model}(a), in the first learning session, the student model $S^{1}$ is derived from the ancestor teacher model $T^{0}$. In the subsequent learning sessions, $S^{m}$ is obtained from
the father teacher model $T^{m}$ via adaptation based on the support set $S'$. The 
adaptation process is formulated as
\begin{equation}
S^{m}=T^{m}({S'}).
\end{equation}

The same as the event detection process of the teacher models, the
probability distribution $\boldsymbol{p}_{stu}$ of the student model
to different label sequences for the query tokens is calculated as
\begin{equation}
\boldsymbol{p}_{stu}=S^{m}({S'},{Q'}).
\end{equation}

\subsubsection{The Loss Functions}

Since Prompt-KD adopts the knowledge distillation framework, its loss functions consist of two parts, i.e., distillation loss and student loss. The distillation loss $l_{dis}$ is obtained upon the cross entropy
loss function $L(.,.)$ between the the probability distributions
$\boldsymbol{p}_{tea}$ and $\boldsymbol{p}_{stu}$ as 
\begin{equation}
l_{dis}=L(\boldsymbol{p}_{tea},\boldsymbol{p}_{stu}).
\end{equation}

In the meantime, the student loss $l_{stu}$ is similarly calculated via
$L(.,.)$ as 
\begin{equation}
l_{stu}=L(\boldsymbol{p}_{stu},\boldsymbol{y}).
\end{equation}

Then, the final loss $l$ is obtained via summing the above two losses:
\begin{equation}
l=l_{dis}+l_{stu}.
\end{equation}

\subsection{Prompt Learning}

To cope with the few-shot learning scenario and alleviate the corresponding new class overfitting problem, Prompt-KD adopts prompt learning as it can concatenate instructions with semantic information after the support instances and thus reduces the dependence on labeled instances. Specifically, Prompt-KD presents a new cloze prompt 
for joint FSED, which is designed as ``This
is a {[}mask{]} event.{[}SEP{]} Its trigger words are {[}mask{]}.''.

Moreover, this module is equipped with a prompt-oriented
curriculum learning mechanism. To the best of our knowledge, this
paper is the first to combine curriculum learning with prompt learning.
In detail, a three-stage curriculum learning based prompt is adopted,
as shown in Table~\ref{tab:cl-prompt}. At Stage 2, the candidate set of
{[}mask{*}{]} is \{``before'', ``now''\}, where ``before''
indicates that the event class has been learned in the past, while ``now''
denotes the event class is being learned at present. At Stage 3, \{``before'',
``recently'', ``now''\} is employed as the candidate set, where
``before'' denotes that the event class has been learned a long time
ago, ``recently'' suggests that the event class has been learned a while
ago, and ``now'' is the same as that at Stage 2.

In practice, dealing with a CIFSED task with $M$ learning sessions, Stages
1-3 contain the sessions 1 to $\lceil\frac{1}{3}M\rceil$, $\lceil\frac{1}{3}M+1\rceil$ to $\lceil\frac{2}{3}M\rceil$, and $\lceil\frac{2}{3}M+1\rceil$ to $M$, respectively, where $\lceil\cdot\rceil$ denotes rounding up to an integer.

\section{Experiments}

\subsection{Datasets and Evaluation Metrics}

\noindent We conduct experiments on two FSED benchmark datasets, i.e.,
FewEvent~\cite{deng2020meta} and MAVEN~\cite{wang2020maven}, which contain 100 and 168 classes, respectively.

\textbf{FewEvent}. FewEvent is designed as a benchmark FSED dataset,
which extracts event classes in ACE-2005~\cite{doddington2004automatic}
and TAC-KBP-2017~\cite{ji2011knowledge}. Besides, it also extends
many new event classes from Wikipedia and Freebase via automatic tagging.
The dataset contains a total of 70,852 instances, with 19 event classes
subdivided into 100 subclasses, where each subclass has an average of
about 700 instances.

\textbf{MAVEN}. MAVEN is a large-scale common domain event detection
dataset that contains 4480 documents and 118732 event instances covering
168 event classes. 
For MAVEN, we adopt 100 classes that have more
than 200 instances, following the previous work~\cite{zhao2022knowledge}. 

We set up four configurations, namely, 5-way 1-shot, 5-way 3-shot,
10-way 1-shot and 10-way 3-shot, for each CIFSED task.
We follow the evaluation protocols in~\cite{tao2020few,dong2021few}
and thus redivide the above two datasets. For both FewEvent and MAVEN, 50 classes are randomly selected as base classes for
$D^{(0)}$ and the other 50 classes are equally split for class-incremental
learning. For the 5-way tasks, the 50 classes especially for class-incremental
learning are equally divided into 10 subsets. Therefore, we obtain
11 subsets (i.e., ${D^{(0)},D^{(1)},...,D^{(10)}}$) totally, where
each $D^{(m)} (m>0)$ has 5 classes and each class contains randomly
sampled 1 or 3 support instances and 1 query instances. Similar
with the 5-way tasks, 6 subsets (i.e., ${D^{(0)},D^{(1)},...,D^{(5)}}$)
are obtained for the 10-way tasks and each $D^{(m)}(m>0)$ contains
10 classes. 

In addition,
we adopt the standard micro F1 score as the evaluation metric and
report the averages upon 5 randomly initialized runs.

\begin{table*}
\centering
\resizebox{2\columnwidth}{!}{
\begin{tabular}{c|cccccccccccc}
\hline 
\multicolumn{13}{c}{{\small{}Dataset: FewEvent}}\tabularnewline
\hline 
\multirow{2}{*}{{\small{}Method}} & \multicolumn{11}{c}{{\small{} Learning Sessions}} & {\small{}Average}\tabularnewline
\cline{2-12} \cline{3-12} \cline{4-12} \cline{5-12} \cline{6-12} \cline{7-12} \cline{8-12} \cline{9-12} \cline{10-12} \cline{11-12} \cline{12-12} 
 & {\small{}0} & {\small{}1} & {\small{}2} & {\small{}3} & {\small{}4} & {\small{}5} & {\small{}6} & {\small{}7} & {\small{}8} & {\small{}9} & {\small{}10} & {\small{}F1}\tabularnewline
\hline 
{\small{}PA-CRF-Meta} & {\small{}78.19} & {\small{}60.67} & {\small{}58.80} & {\small{}56.25} & {\small{}55.82} & {\small{}54.72} & {\small{}53.82} & {\small{}52.67} & {\small{}51.82} & {\small{}46.62} & {\small{}44.78} & {\small{}55.83}\tabularnewline
\hline 
{\small{}PA-CRF-CIL} & {\small{}78.19} & {\small{}55.95} & {\small{}47.36} & {\small{}45.90} & {\small{}43.03} & {\small{}41.36} & {\small{}39.78} & {\small{}37.78} & {\small{}36.79} & {\small{}35.09} & {\small{}32.78} & {\small{}44.91}\tabularnewline
\hline 
{\small{}Prompt-KD} & {\small{}78.19} & {\small{}60.58} & {\small{}51.78} & {\small{}50.31} & {\small{}49.69} & {\small{}50.53} & {\small{}48.61} & {\small{}47.57} & {\small{}47.32} & {\small{}47.03} & {\small{}45.22} & {\small{}52.44}\tabularnewline
\hline 
\hline 
\multicolumn{13}{c}{{\small{}Dataset: MAVEN}}\tabularnewline
\hline 
\multirow{2}{*}{{\small{}Method}} & \multicolumn{11}{c}{{\small{}Learning Sessions}} & {\small{}Average}\tabularnewline
\cline{2-12} \cline{3-12} \cline{4-12} \cline{5-12} \cline{6-12} \cline{7-12} \cline{8-12} \cline{9-12} \cline{10-12} \cline{11-12} \cline{12-12} 
 & {\small{}0} & {\small{}1} & {\small{}2} & {\small{}3} & {\small{}4} & {\small{}5} & {\small{}6} & {\small{}7} & {\small{}8} & {\small{}9} & {\small{}10} & {\small{}F1}\tabularnewline
\hline 
{\small{}PA-CRF-Meta} & {\small{}73.04} & {\small{}42.83} & {\small{}38.97} & {\small{}37.92} & {\small{}35.66} & {\small{}33.33} & {\small{}32.72} & {\small{}29.83} & {\small{}29.51} & {\small{}26.05} & {\small{}25.46} & {\small{}36.84}\tabularnewline
\hline 
{\small{}PA-CRF-CIL} & {\small{}73.04} & {\small{}35.02} & {\small{}30.93} & {\small{}28.60} & {\small{}26.23} & {\small{}25.00} & {\small{}24.14} & {\small{}23.07} & {\small{}22.61} & {\small{}21.68} & {\small{}18.96} & {\small{}29.93}\tabularnewline
\hline 
{\small{}Prompt-KD} & {\small{}73.04} & {\small{}39.47} & {\small{}36.98} & {\small{}35.07} & {\small{}32.98} & {\small{}30.03} & {\small{}30.05} & {\small{}28.02} & {\small{}27.15} & {\small{}25.89} & {\small{}23.71} & {\small{}34.76}\tabularnewline
\hline 
\end{tabular}
}
\caption{The F1 scores (\%) of the 5-way 1-shot tasks on two benchmark datasets: FewEvent and MAVEN. }

\label{tab:main experiment 5w1s}
\end{table*}

\begin{table*}
\centering {\small{}}%
\begin{tabular}{c|ccccccc}
\hline 
\multicolumn{8}{c}{{\small{}Dataset: FewEvent}}\tabularnewline
\hline 
\multirow{2}{*}{{\small{}Method}} & \multicolumn{6}{c}{{\small{}Learning Sessions}} & {\small{}Average}\tabularnewline
\cline{2-7} \cline{3-7} \cline{4-7} \cline{5-7} \cline{6-7} \cline{7-7} 
 & {\small{}0} & {\small{}1} & {\small{}2} & {\small{}3} & {\small{}4} & {\small{}5} & {\small{}F1}\tabularnewline
\hline
{\small{}PA-CRF-Meta} & {\small{}77.84} & {\small{}61.95} & {\small{}58.23} & {\small{}54.88} & {\small{}54.58} & {\small{}54.13} & {\small{}60.26}\tabularnewline
\hline 
{\small{}PA-CRF-CIL} & {\small{}77.84} & {\small{}50.17} & {\small{}44.49} & {\small{}33.04} & {\small{}28.88} & {\small{}23.31} & {\small{}42.95}\tabularnewline
\hline 
{\small{}Prompt-KD} & {\small{}77.84} & {\small{}55.48} & {\small{}47.69} & {\small{}40.79} & {\small{}34.09} & {\small{}32.05} & {\small{}47.99}\tabularnewline
\hline 
\hline 
\multicolumn{8}{c}{{\small{}Dataset: MAVEN}}\tabularnewline
\hline 
\multirow{2}{*}{{\small{}Method}} & \multicolumn{6}{c}{{\small{}Learning Sessions}} & {\small{}Average}\tabularnewline
\cline{2-7} \cline{3-7} \cline{4-7} \cline{5-7} \cline{6-7} \cline{7-7} 
 & {\small{}0} & {\small{}1} & {\small{}2} & {\small{}3} & {\small{}4} & {\small{}5} & {\small{}F1}\tabularnewline
\hline 
{\small{}PA-CRF-Meta} & {\small{}69.04} & {\small{}31.58} & {\small{}26.26} & {\small{}24.94} & {\small{}20.68} & {\small{}19.38} & {\small{}31.98}\tabularnewline
\hline 
{\small{}PA-CRF-CIL} & {\small{}69.04} & {\small{}26.85} & {\small{}21.17} & {\small{}16.67} & {\small{}14.47} & {\small{}11.69} & {\small{}26.64}\tabularnewline
\hline 
{\small{}Prompt-KD} & {\small{}69.04} & {\small{}28.40} & {\small{}24.07} & {\small{}19.78} & {\small{}18.23} & {\small{}16.95} & {\small{}30.88}\tabularnewline
\hline 
\end{tabular}{\small\par}

\caption{The F1 scores (\%) of the 10-way 1-shot tasks on two datasets: FewEvent and MAVEN. }

\label{tab:main experiments 10w1s}
\end{table*}

\begin{table*}[t]
\centering 
\begin{tabular}{c|ccccccccccc}
\hline 
\multirow{2}{*}{{\small{}Method}} & \multirow{2}{*}{{\small{}KD}} & \multirow{2}{*}{{\small{}AT}} & \multirow{2}{*}{{\small{}ATT}} & \multirow{2}{*}{{\small{}PL}} & \multirow{2}{*}{{\small{}CL}} & \multicolumn{5}{c}{{\small{}Learning Sessions}} & {\small{}Average}\tabularnewline
\cline{7-11} \cline{8-11} \cline{9-11} \cline{10-11} \cline{11-11} 
 &  &  &  &  &  & {\small{}1} & {\small{}2} & {\small{}3} & {\small{}4} & {\small{}5} & {\small{}F1}\tabularnewline
\hline 
{\small{}Prompt-KD} & $\boldsymbol{\surd}$ & $\boldsymbol{\surd}$ & $\boldsymbol{\surd}$ & $\boldsymbol{\surd}$ & $\boldsymbol{\surd}$ & \textbf{\small{}55.48} & \textbf{\small{}47.69} & \textbf{\small{}40.79} & \textbf{\small{}34.09} & \textbf{\small{}32.05} & \textbf{\small{}42.02}\tabularnewline
\hline 
$\mathcal{A}$ & $\boldsymbol{\surd}$ & $\boldsymbol{\surd}$ & $\boldsymbol{\surd}$ & $\boldsymbol{\surd}$ &  & {\small{}55.48} & {\small{}47.34} & {\small{}40.33} & {\small{}33.79} & {\small{}31.51} & {\small{}41.69}\tabularnewline
$\mathcal{B}$ & $\boldsymbol{\surd}$ & $\boldsymbol{\surd}$ & $\boldsymbol{\surd}$ &  &  & {\small{}54.36} & {\small{}46.12} & {\small{}38.02} & {\small{}30.19} & {\small{}27.77} & {\small{}39.29}\tabularnewline
$\mathcal{C}$ & $\boldsymbol{\surd}$ & $\boldsymbol{\surd}$ &  &  &  & {\small{}51.64} & {\small{}45.98} & {\small{}37.49} & {\small{}30.01} & {\small{}26.93} & {\small{}38.41}\tabularnewline
$\mathcal{D}$ & $\boldsymbol{\surd}$ &  &  &  &  & {\small{}51.64} & {\small{}45.38} & {\small{}34.60} & {\small{}29.57} & {\small{}24.94} & {\small{}37.23}\tabularnewline
\hline 
{\small{}PA-CRF-CIL} &  &  &  &  &  & {\small{}50.17} & {\small{}44.49} & {\small{}33.04} & {\small{}28.88} & {\small{}23.31} & {\small{}35.97}\tabularnewline
\hline 
\end{tabular}
\caption{The results of the ablation study on the 10-way 1-shot tasks on FewEvent.}
\label{tab:ablation study}
\end{table*}

\subsection{Implementation Details and Parameter Setting}

BERT-base-uncased
\cite{kenton2019bert} is employed as the encoder for both the teacher models and the student model, whose input
sentence has max length of 128 and the hidden size $H$ is 768. Prompt-KD is trained with the 1e-5 learning rate with the AdamW optimizer.
Moreover, the dropout is 0.1 and the batch size is 1. We pre-train PA-CRF with 10,000 episodes
on $D^{(0)}$ as the ancestor teacher model. Furthermore, we evaluate
the performance on $D^{(0)}\cup D^{(1)}\cup...\cup D^{(m)}$ in the $m$-th learning session, all
following the episodic paradigm. We run all experiments using PyTorch
1.5.1 on the Nvidia V100 GPU with 32GB memory, Intel(R) Xeon(R) Gold
5218 CPU @ 2.30GHz with 128GB memory on CentOS Linux release 7.9.2009
(Core).

\subsection{Baseline Models}

Since we are the first to propose the CIFSED task, there are no existing methods for it. In order to investigate the validity and effectiveness of Prompt-KD, we develop two variants (i.e., PA-CRF-CIL and PA-CRF-Meta) of the first and representative joint FSED method, i.e., PA-CRF, for experimental comparison, as it is the base model of Prompt-KD. Specifically, PA-CRF-CIL is obtained by applying the PA-CRF model in the class-incremental scenario, where the model in the $m$-th learning session is derived via fine-tuning the model from the last learning session based on $D^{(m)}$ and the corresponding exemplars.
PA-CRF-Meta is trained via episodic learning under the meta learning framework on the joint dataset $D^{(0)}\cup D^{(1)}\cup...\cup D^{(m)}$, which can thus be regarded, to a certain degree, as an upper bound model.

\subsection{Experimental Results}

\begin{figure}[t]
\centering \includegraphics[width=0.5\textwidth]{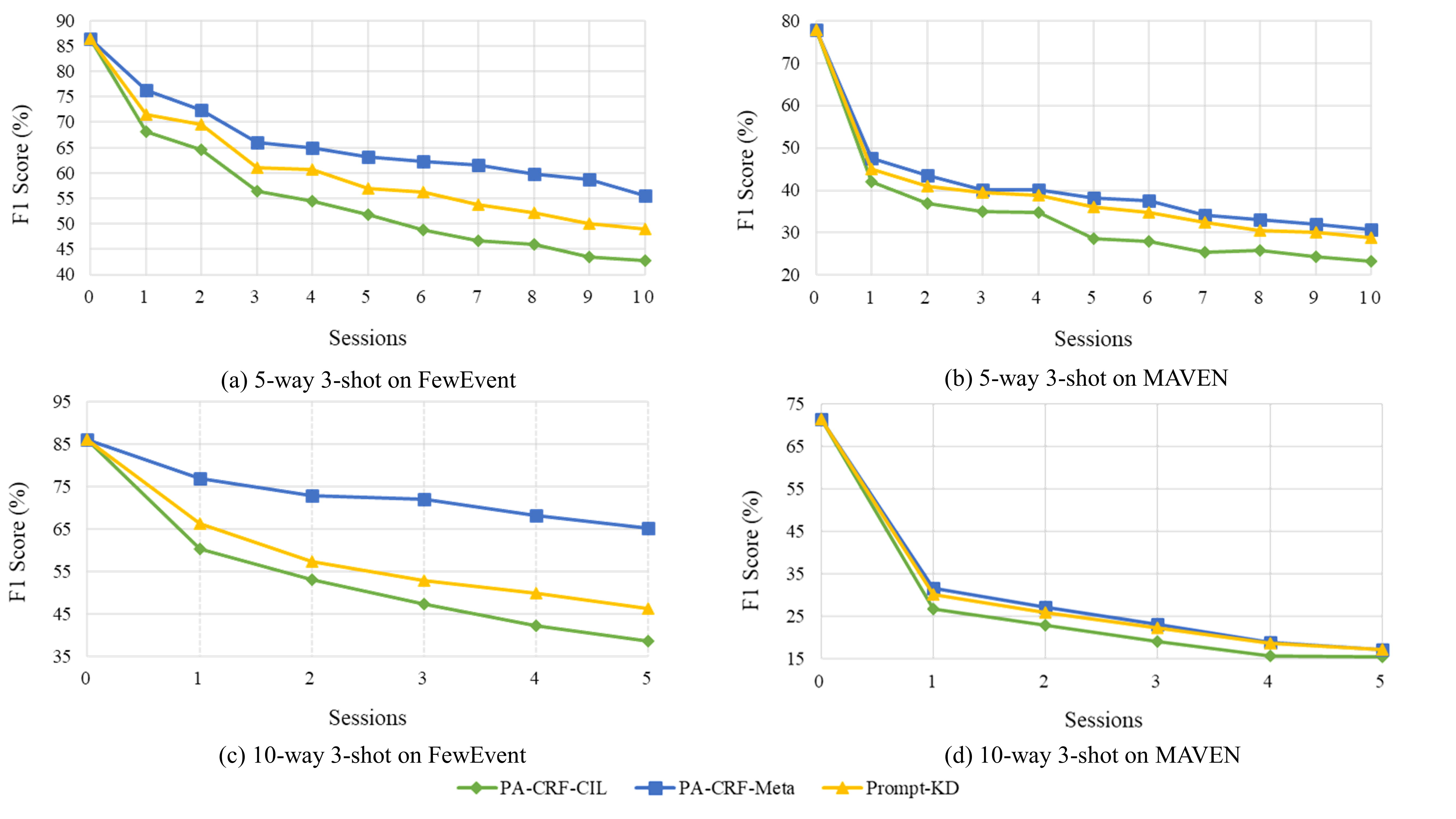}
\caption{The F1 score curves of the 5-way 3-shot and the 10-way 3-shot tasks on FewEvent
and MAVEN.}
\label{fig:curve}
\end{figure}

Tables~\ref{tab:main experiment 5w1s} and~\ref{tab:main experiments 10w1s} present the overall experimental
results on the 5-way 1-shot and the 10-way 1-shot tasks, whilst Figure~\ref{fig:curve}
compares the test F1 scores of the 5-way 3-shot and the 10-way 3-shot tasks
on FewEvent and MAVEN, respectively. We summarize the results
as follows.
\begin{itemize}
\item Our Prompt-KD method outperforms the PA-CRF-CIL baseline and achieves
the state-of-the-art performance consistently on both datasets, all tasks and all learning sessions.
Particularly, the average F1 score of Prompt-KD increases by 5-8\%
on FewEvent and 4-5\% on MAVEN, respectively, compared to PA-CRF-CIL.

\item As shown in Figure~\ref{fig:curve}, the F1 score curve of Prompt-KD
is close to that of PA-CRF-Meta on MAVEN, where these 
curves seem almost overlapping. However, there exists a clear distance
between the F1 score curves of Prompt-KD and PA-CRF-Meta on FewEvent,
which indicates that the CIFSED method has great potential to be improved
on FewEvent.

\item The average F1 scores of Prompt-KD and PA-CRF-CIL on FewEvent on the 5-way tasks are higher than those on the 10-way tasks, as shown in Tables~\ref{tab:main experiment 5w1s} and~\ref{tab:main experiments 10w1s} as well as Figure~\ref{fig:curve}. This phenomenon indicates that learning with fewer learning sessions with more classes suffers from a more serious old knowledge forgetting problem than that with more learning sessions with fewer classes.

\item As shown in Table~\ref{tab:main experiment 5w1s}, in Sessions 9 and 10 on the 5-way
1-shot tasks on FewEvent, our Prompt-KD method even outperforms PA-CRF-Meta, which can further demonstrate the 
ability of Prompt-KD to overcome the old knowledge forgetting problem.
This phenomenon may be due to that PA-CRF-Meta is not a theoretical upper bound model, where although the meta learning framework can alleviate the class imbalance problem, it also damages the performance of the PA-CRF-Meta model on base classes with a large number of instances.
 
\end{itemize}

\subsection{Ablation Study}

We conduct ablation studies to investigate the
effectiveness of Knowledge Distillation (KD), Ancestor Teacher
(AT), Attention (ATT), Prompt Learning (PL) and Curriculum Learning
(CL), as well as their impacts on the performance of Prompt-KD on
the 10-way 1-shot tasks. Without loss of generality, these ablation studies are carried out on FewEvent. Specifically, the ablated models
of Prompt-KD incrementally without CL, (PL and CL), (ATT, PL and CL), (AT, ATT, PL
and CL) are identified as $\mathcal{A}$, $\mathcal{B}$, $\mathcal{C}$ and $\mathcal{D}$,
respectively. As shown in Table~\ref{tab:ablation study}, the performance of
the ablated model $\mathcal{A}$ falls compared to that of Prompt-KD,
which indicates that CL contributes to the effectiveness of Prompt-KD.
Similarly, the comparisons between the results of the ablated models $\mathcal{A}$ and
$\mathcal{B}$, $\mathcal{B}$ and $\mathcal{C}$, $\mathcal{C}$ and $\mathcal{D}$,
as well as $\mathcal{D}$ and PA-CRF-CIL, demonstrate the effectiveness of PL,
ATT, AT and KD, respectively.

\subsection{In-depth Analysis}

In this subsection, we compare Prompt-KD with other class-incremental event detection methods to verify its effectiveness on the method level. Subsequently, we conduct experiments to demonstrate the contributions of ATT, AT and PL, which are essential to the performance of Prompt-KD.

\begin{table*}
\centering
\small
\begin{tabular}{ccccccc}
\hline 
\multicolumn{7}{c}{{Dataset: MAVEN}}\tabularnewline
\hline 
\multirow{2}{*}{{Method}} & \multicolumn{5}{c}{{Learning Sessions}} & {Average}\tabularnewline
\cline{2-6} 
& {1} & {2} & {3} & {4} & {5} & {F1}\tabularnewline
\hline
{KDR} & {17.99} & {15.02} & {10.56} & {9.07} & {6.37} & {11.80}\tabularnewline
\hline 
{EMP} & {18.09} & {16.03} & {10.93} & {8.94} & {6.43} & {12.08} \tabularnewline
\hline 
{PA-CRF-CIL} & {26.85} & {21.17} & {16.67} & {14.47} & {11.69} & {26.64}\tabularnewline
\hline 
{Prompt-KD} & {28.40} & {24.07} & {19.78} & {18.23} & {16.95} & {30.88}\tabularnewline
\hline 
\end{tabular}
\caption{Comparison results with class-incremental event detection methods.}
\label{tab:with emp and kdr}
\end{table*}

\begin{table}
\centering
\resizebox{1\columnwidth}{!}{
\begin{tabular}{cccccccc}
\hline 
\multicolumn{8}{c}{{Dataset: FewEvent}}\tabularnewline
\hline 
\multirow{2}{*}{{Method}} & \multirow{2}{*}{{\#Instances}} & \multicolumn{6}{c}{{Learning Sessions}}\tabularnewline
\cline{3-8}
 & {} & {0} & {1} & {2} & {3} & {4} & {5}\tabularnewline
\hline
{$\mathcal{C}$} & {50} & {39} & {31} & {25} & {21} & {16} & {12} \tabularnewline
\hline 
{$\mathcal{D}$} & {50} & {39} & {27} & {22} & {17} & {11} & {7}\tabularnewline
\hline 
\end{tabular}
}
\caption{\quad The case study of the ablated models $\mathcal{C}$ and $\mathcal{D}$}
\label{tab:50_cases}
\end{table}

\begin{table*}[t]
\centering%
\begin{tabular}{>{\raggedright}p{9cm}ccccccc}
\hline 
\multirow{2}{9cm}{\centering{}{\small{}Instances}} & \multirow{2}{*}{{\small{}Method}} & \multicolumn{6}{c}{{\small{}Learning Sessions}}\tabularnewline
\cline{3-8} \cline{4-8} \cline{5-8} \cline{6-8} \cline{7-8} \cline{8-8} 
 &  & {\small{}0} & {\small{}1} & {\small{}2} & {\small{}3} & {\small{}4} & {\small{}5}\tabularnewline
\hline 
\multirow{2}{9cm}{{\small{}Denis Rancourt is a former professor (}\textit{\textcolor{blue}{\small{}B-Education.Education}}{\small{})
of physics at the University of Ottawa.}} & $\mathcal{C}$ & \textcolor{teal}{\small{}$\boldsymbol{\surd}$} & \textcolor{teal}{\small{}$\boldsymbol{\surd}$} & \textcolor{teal}{\small{}$\boldsymbol{\surd}$} & \textcolor{teal}{\small{}$\boldsymbol{\surd}$} & \textcolor{teal}{\small{}$\boldsymbol{\surd}$} & \textcolor{teal}{\small{}$\boldsymbol{\surd}$}\tabularnewline
\cline{2-8} \cline{3-8} \cline{4-8} \cline{5-8} \cline{6-8} \cline{7-8} \cline{8-8} 
 & $\mathcal{D}$ & \textcolor{teal}{\small{}$\boldsymbol{\surd}$} & \textcolor{teal}{\small{}$\boldsymbol{\surd}$} & \textcolor{teal}{\small{}$\boldsymbol{\surd}$} & \textcolor{red}{$\boldsymbol{\times}$} & \textcolor{red}{$\boldsymbol{\times}$} & \textcolor{red}{$\boldsymbol{\times}$}\tabularnewline
\hline 
\multirow{2}{9cm}{{\small{}He says that 20 \% of the people who get that card send (}\textit{\textcolor{blue}{\small{}B-Contact.E-mail}}{\small{})
him an e-mail.}} & $\mathcal{C}$ & \textcolor{teal}{\small{}$\boldsymbol{\surd}$} & \textcolor{teal}{\small{}$\boldsymbol{\surd}$} & \textcolor{teal}{\small{}$\boldsymbol{\surd}$} & \textcolor{teal}{\small{}$\boldsymbol{\surd}$} & \textcolor{teal}{\small{}$\boldsymbol{\surd}$} & \textcolor{teal}{\small{}$\boldsymbol{\surd}$}\tabularnewline
\cline{2-8} \cline{3-8} \cline{4-8} \cline{5-8} \cline{6-8} \cline{7-8} \cline{8-8} 
 & $\mathcal{D}$ & \textcolor{teal}{\small{}$\boldsymbol{\surd}$} & \textcolor{teal}{\small{}$\boldsymbol{\surd}$} & \textcolor{red}{$\boldsymbol{\times}$} & \textcolor{red}{$\boldsymbol{\times}$} & \textcolor{red}{$\boldsymbol{\times}$} & \textcolor{red}{$\boldsymbol{\times}$}\tabularnewline
\hline 
\end{tabular}

\caption{The case study of the ablated models $\mathcal{C}$ and $\mathcal{D}$ on the 10-way
1-shot tasks on FewEvent. The blue words denote the ground truth labels,
the green check marks indicate that the model provides correct answers,
whilst the red cross marks suggest that the model makes wrong predictions.}

\label{tab:case study}
\end{table*}

\subsubsection{Comparison between Prompt-KD and Class-Incremental Event Detection Methods}

To further demonstrate the effectiveness of Prompt-KD, we compare it with two class-incremental event detection methods, i.e., EMP~\cite{liu2022incremental} and KDR~\cite{yu2021lifelong}. EMP is the state-of-the-art method on MAVEN for class-incremental event detection, and KDR is the second state-of-the-art method. The experimental results of 10-way 1-shot tasks on MAVEN are shown in Table~\ref{tab:with emp and kdr}. As we can see, EMP and KDR achieve worse performance than Prompt-KD, even worse than PA-CRF-CIL. This may be due to that these methods cannot automatically adapt to the few-shot scenarios.

\subsubsection{Visualization Analysis of the Attention Mechanism}

\begin{figure}[t]
\centering \includegraphics[width=0.5\textwidth]{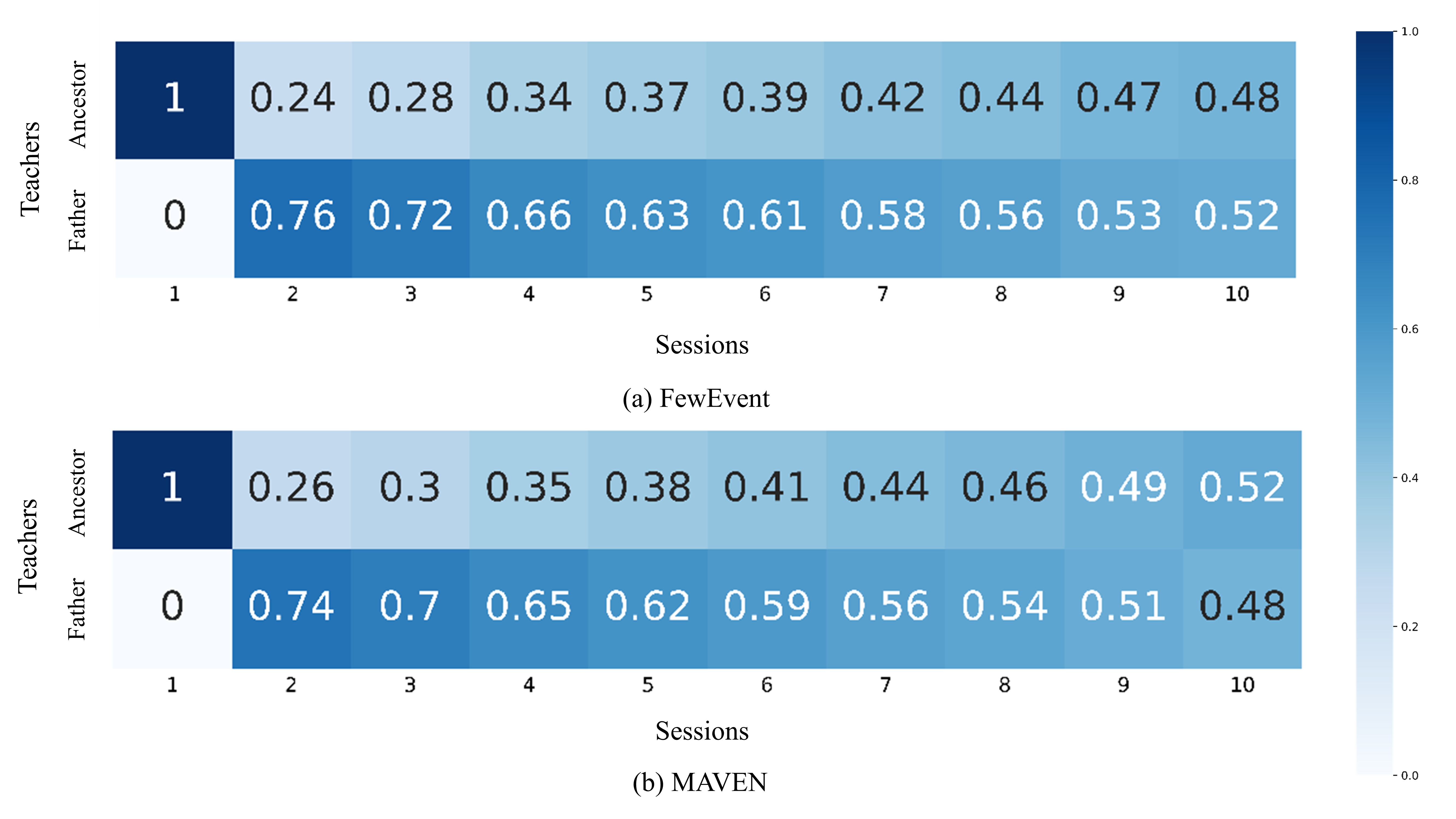}
\caption{The heat map of different weights of the teacher models on the 5-way 1-shot tasks on FewEvent (a)
and MAVEN (b).}
 \label{fig:visualization}
\end{figure}

To further demonstrate the effectiveness of the attention mechanism, we draw a heat map on the
5-way 1-shot tasks on FewEvent and MAVEN to visualize the different
weights of the ancestor and father teacher models. As shown in
Figure~\ref{fig:visualization}, the weight of the father teacher model
far exceeds that of the ancestor teacher model in the first few learning sessions.
However, with new classes arriving constantly, the weight of the
ancestor teacher model gradually increases on both two datasets and even exceeds that of the
father teacher model on MAVEN. This
situation is in line with the intuitive cognition that, the model
forgets more knowledge about base classes as time goes by. Therefore,
the model needs to assign higher weight to the ancestor teacher model to
review the base knowledge.

\subsubsection{Case Studies on the Ancestor Teacher Model}

To illustrate the contributions of the ancestor teacher model, we conduct experiments on the 10-way 1-shot tasks on FewEvent between the ablated model $\mathcal{C}$ with AT and the ablated model $\mathcal{D}$ without AT. We choose 50 instances (1-shot for 50 base classes) and count how many of them are correctly predicted in the subsequent learning sessions. As shown in Table~\ref{tab:50_cases}, as the learning session goes by, the model $\mathcal{C}$ predicts more correct instances than the model $\mathcal{D}$. It indicates that the model with AT is less likely to forget the base knowledge, demonstrating the effectiveness of AT in alleviating the old knowledge forgetting problem.

Specifically, we choose two instances of classes \textit{Education.Education}
and \textit{Contact.E-Mail} from $D^{(0)}$ of FewEvent 
 and present their prediction results of the models
$\mathcal{C}$ and $\mathcal{D}$. As shown in Table
\ref{tab:case study}, the ablated model $\mathcal{C}$ correctly makes predictions
on both two instances and in all learning sessions. Nevertheless, the
model $\mathcal{D}$ provides wrong answers of the first instance
in Sessions 3-5 and of the second one in Sessions 2-5, respectively.
This may be because that the model $\mathcal{D}$ puts more attention on the
new classes and thus begins to forget base classes after 2-3 learning sessions.
The above phenomena can demonstrate that, reusing
the ancestor teacher model constantly can effectively
overcome the forgetting problem about base knowledge.

\subsubsection{Experimental Analysis on Prompt Learning}

To verify the effectiveness of prompt learning for coping with the few-shot scenarios and alleviating the corresponding new class overfitting problem, we conduct experiments on the 10-way 1-shot tasks on FewEvent comparing the F1 difference of the ablated method $\mathcal{A}$ with prompt learning and the ablated method $\mathcal{B}$ without prompt learning on the support set and the query set. The experimental results are shown in Table~\ref{tab:prompt_learning}, where we can see that the difference of support set and query set decreases and the performance on the query set increases of the model $\mathcal{A}$ comparing with the model $\mathcal{B}$. It indicates that prompt learning can alleviate the new class overfitting problem.

\begin{table*}
\small
\centering
\begin{tabular}{cccccccc}
\hline 
\multicolumn{8}{c}{{Dataset: FewEvent}}\tabularnewline
\hline 
\multirow{2}{*}{{Method}} & \multirow{2}{*}{{Dataset}} & \multicolumn{5}{c}{{Learning Sessions}} & {Average}\tabularnewline
\cline{3-7}
 & {} & {1} & {2} & {3} & {4} & {5} & {F1}\tabularnewline
\hline
{$\mathcal{A}$} & {Support Set} & {67.25} & {58.41} & {51.33} & {43.22} & {40.78} & {52.19} \tabularnewline
\hline 
{$\mathcal{A}$} & {Query Set} & {55.48} & {47.34} & {40.33} & {33.79} & {31.51} & {41.69}\tabularnewline
\hline 
{$\mathcal{B}$} & {Support Set} & {67.03} & {58.10} & {51.07} & {42.96} & {40.27} & {51.88} \tabularnewline
\hline 
{$\mathcal{B}$} & {Query Set} & {54.36} & {46.12} & {38.02} & {30.19} & {27.22} & {39.29}\tabularnewline
\hline 
\end{tabular}
\caption{The F1 scores (\%) of the models $\mathcal{A}$ and $\mathcal{B}$ on the support and query set.}
\label{tab:prompt_learning}
\end{table*}

\section{Conclusions and Future Work}

In this paper, we proposed a new task, i.e., CIFSED, and a
knowledge distillation and prompt learning based method, called Prompt-KD,
for it. Specifically, to overcome the old knowledge forgetting problem, Prompt-KD develops an attention based multi-teacher knowledge distillation framework, where the ancestor teacher model pre-trained on base classes is reused in all learning sessions. 
Moreover, to cope with the few-shot scenario and alleviate the corresponding new class overfitting problem, Prompt-KD is equipped with
a prompt learning mechanism. Extensive experiments on
two benchmark datasets, i.e., FewEvent and MAVEN, demonstrate the
superior performance of Prompt-KD. 

\section{Acknowledgments}

The work is supported by the National Key Research and Development Project of China, the GFKJ Innovation Project, the Beijing Academy of Artificial intelligence under grant BAAl2019ZD0306, the KGJ Project under grant JCKY2022130C039, and the Lenovo-CAS Joint Lab Youth Scientist Project. We appreciate anonymous reviewers for their insightful comments and suggestions.

\section{Bibliographical References}

\bibliographystyle{lrec-coling2024-natbib}
\bibliography{reference}

\end{document}